# A Locally Deployed RAG-Based Academic Advising System for Course Selection

Feng Li[a], Yoritaka Iwata[b]*

[a]*Shiga University, 1-1-1 banba, Hikone, Shiga 522-8522, Japan*
[b]*Osaka University of Economics and Law, 2-10-45 Kitahonmachi, Yao, Osaka 581-8522, Japan*

**Abstract**

The correct sequence of courses in the curriculum based on prerequisites between courses is of great importance for students to develop their knowledge and skills holistically. However, students crafting this sequence in isolation frequently struggle with recognition limitations and information overload that leads to confusion. Simultaneously, education institutions encounter difficulties in providing adequate academic advice for the correct sequence due to limited education resources. To address these challenges, we propose a locally deployed RAG-based academic advising system grounded in syllabus information. By combining large language models with retrieval from structured syllabus data, the system is designed to support course selection, prerequisite understanding, and personalized study planning in a privacy-preserving manner.

*Keywords:* Retrieval-Augmented Generation (RAG); academic advising; course selection; large language models; privacy-preserving AI.

## 1. Introduction

Academic advising for course registration plays a crucial role in leading students to be successful. However, it requires education institutions to heavily invest in training professional advisors who give one-on-one consultations. Moreover, those human advisors often struggle with hurdles like overwhelming caseloads, knowledge management, the potential for giving inaccurate advice, and disengagement [1]. Those can hinder students from consulting with their advisors. Otherwise, while a full list of courses together with their descriptions in syllabus allows students to know everything, searching complicated information on syllabus autonomously and information overload can frustrate students. These factors contribute to selecting courses simply based on the title of courses rather than comprehensive research. Moreover, while the curriculum provides students with static and extensive information through syllabuses to help students select courses effectively, hindering personalized and diverse learning paths and data-driven decision making.

To address the restrictions in human advisors, AI-powered chatbot has significant potential advantages to assist students for advisors, providing students with information and resources needed to establish degree plans aligned with their academic goals.

In recent years, AI has emerged as a key driver in reconfiguring world across diverse sectors. Resent research explored adoption of AI in academic advising landscapes to elevate institutional efficiency and student outcomes [2,3]. AI-powered chatbots offer significant potential to enhance student advising in higher education by providing tailored guidance. While AI-powered chatbot is revolutionizing the way information is provided, adopting AI is both a significant opportunity and a formidable challenge. Generative AI is unable to handle queries which require dedicated and up-to-date information among institutions. Otherwise, training new dedicated model for chatbot requires data and computational resources that are out of the reach of most people and organizations [7].

* Corresponding author.
*E-mail address:* y-iwata@s.keiho-u.ac.jp (Yoritaka Iwata)

Our research aims to address those challenges by developing Syllabot, a locally deployed RAG-based academic advising system that can reduce information retrieval costs and facilitate course selection using syllabus information. Furthermore, this study evaluates retrieval performance, answer generation quality, and refusal behavior across unanswerable queries. The main novelty and contributions of this study are summarized as follows:

- Proposed a retrieval-oriented query taxonomy based on the number of evidence chunks required to answer a query.
- Proposed an automated method for generating evaluation queries with gold chunk labels from predefined chunks.
- Implemented a locally deployed system to reduce the risk of sensitive information exposure.
- Demonstrated that BM25, semantic, and hybrid retrieval strategies have different strengths depending on query retrieval complexity.
- Evaluated the system's boundary capability by gradually increasing the context provided to the LLM.

This paper is organized as follows: Section 2 provides an overview of AI-driven advising chatbot. Section 3 presents the methodology, implementation, and evaluation design of Syllabot. Section 4 reports the experimental results, discusses their implications. Section 5 summarizes the paper and discusses limitations and future work.

## 2. Background

The abundance of course options offered by universities may be particularly appealing to new students who are undecided, yet it may also contribute to confusion and indecision [4]. For students, especially those who are new to university, course selection is not simply a matter of choosing interesting subjects. They must also consider factors such as course sequence, prerequisites, workload, and long-term academic goals. In this sense, bounded rationality can lead students to three potential problems: mismatches, dissatisfaction, and failure to meet the requirements for credit.

This problem is also consistent with findings from choice overload research. Experimental evidence from Iyengar and Lepper [5] indicates that people are more likely to make a choice when they are offered a limited set of options rather than an extensive set. Although their study was conducted in a consumer setting, the underlying implication is relevant to course registration: when students are faced with an excessive number of options and large amounts of information, decision-making can become more difficult rather than more effective.

To address such difficulties, recent work has shown that RAG [6] can be useful in student-support settings. Tamascelli et al. proposed ARGObot [9], a RAG system designed as an adviser that integrates information retrieval from a university student handbook to answer students' queries. This chatbot achieves a notable answer correctness score of 0.815 and a context precision score of 0.883, showing the potential of RAG for student advising. However, it adopted a fixed size chunking strategy, which can be limited when applied to semi-structured documents such as syllabi. Syllabus documents often contain section titles, tables, course metadata, prerequisites, assessment criteria, and textbook information. In this case, fixed-size chunking may split these semantic units and weaken retrieval quality.

Moreover, some students' queries can be easily answered by matching evidence with explicit keywords, while others require semantic paraphrase understanding or the collection of multiple evidence chunks across different syllabus sections. This variation means that the difficulty of retrieval is not uniform across student queries. If retrieval performance is evaluated only by an overall score, the strengths and weaknesses of each retrieval strategy may be hidden. Therefore, syllabus-based RAG systems require an evaluation design that separates queries by retrieval complexity and examines how different retrieval strategies perform under each query type.

## 3. Methodology

### *3.1. System Overview*

This section presents the design and implementation of Syllabot. The system uses syllabus documents as its knowledge source and generates answers grounded in retrieved chunks. As shown in Figure 1, the system consists of four main components: syllabus preprocessing and chunking, retrieval index construction, retrieval modules, and a generation module. First, the documents are converted into header-based chunks enriched with metadata. Second, the chunks are indexed for later retrieval. Third, given a student query, the system retrieves relevant chunks using a specified retrieval strategy. Finally, the selected chunks are used as the LLM input context, and the LLM generates a concise answer grounded only in that context. In this implementation, the generation model is executed locally instead of through an external cloud API.

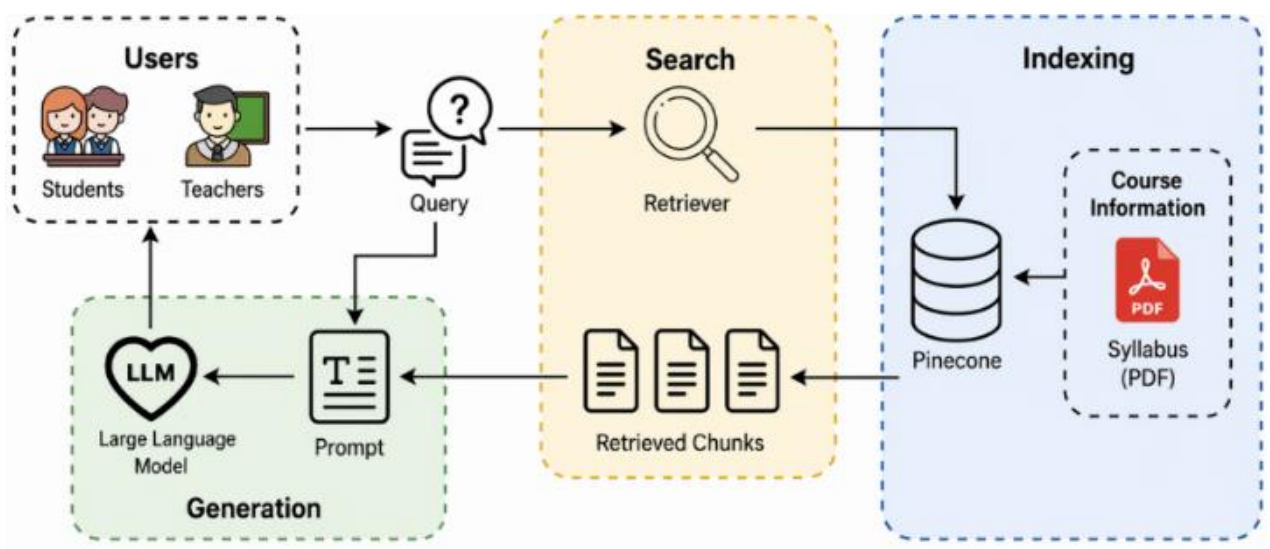

Figure 1. Overall workflow of Syllabot, including syllabus indexing, chunk retrieval, and response generation.

### *3.2. Syllabus Data Collection and Processing*

We collected course information for 97 courses from a Japanese university and converted each document into smaller header-based chunks. As shown in Figure 2, each chunk was annotated with metadata containing concise syllabus information and a source path indicating where the chunk came from.

```
{
  "chunk_id": "EFC2001_chunk_003",
  "course_id": "EFC2001",
  "course_title": "Microeconomics I",
  "instructor": "...",
  "credits": 2,
  "prerequisites": [],
  "section_title": "Course Objectives",
  "source_path": "Microeconomics I > Course Objectives",
  "page_content": "..."
}
```

Figure 2. Example of a metadata-enriched chunk, original language is Japanese.

### *3.3. Retrieval Strategies*

To evaluate how retrieval methods behave across different query types, this study compares three strategies including BM25-based lexical retrieval, embedding-based semantic similarity retrieval, and hybrid retrieval.

#### *3.3.1. BM25-based lexical retrieval*

BM25-based lexical retrieval is configured to evaluate keyword-based matching ability. It is expected to perform well when queries contain explicit terms. This retriever applies Janome for Japanese tokenization and ranks chunks using the BM25 algorithm.

### *3.3.2. Embedding-based Semantic Retrieval*

Embedding-based semantic retrieval is designed to evaluate semantic matching ability. It is expected to excel at dealing with queries that are paraphrased. In this method, chunks are embedded using multilingual-e5-large and stored in Pinecone with their metadata. Given a query, the query is also embedded using the same model, and chunks are ranked according to vector similarity.

### *3.3.3. Hybrid Retrieval*

Hybrid retrieval is introduced to examine whether lexical and semantic retrieval can complement each other. It is expected to balance precision and recall across different queries. In this method, lexical and semantic retrieval are performed independently to obtain candidate chunks. The two ranked lists are then combined using RRF (Reciprocal Rank Fusion) [10]. The final top-k chunks are selected based on the RRF score.

## *3.4. Generation module*

The generation model uses RakutenAI-7B-instruct [11], a locally executed LLM developed for Japanese and English. In this study, the temperature is set to 0.0 and the maximum output length is set to 256 tokens. To keep the generation condition consistent, the same model and the same prompt are used across all retrieval settings.

For each query, selected retrieved chunks are provided to the LLM as input context. The prompt was written in Japanese because the target documents and user queries are Japanese.

## *3.5. Retrieval-oriented Query Taxonomy*

To compare the performance of different retrieval strategies under different retrieval complexity, the evaluation queries are divided into four types based on retrieval complexity, which is defined by both the number of gold chunks required and the degree of lexical overlap between the query and the gold chunks. Gold chunks refer to the syllabus chunks required to answer each query. Each evaluation query is assigned one or more gold chunks, except for Out-of-scope Queries. The four query types are defined as follows:

- Lexical Match Queries are answerable from a single gold chunk and contain keywords explicitly appearing in that chunk.
- Semantic Paraphrase Queries are generated from a single gold chunk and express the same meaning with reduced keyword overlap.
- Multi-evidence Synthesis Queries require multiple gold chunks to answer and are used to evaluate whether the retriever can collect a complete evidence set.
- Out-of-scope Queries are outside the scope of the syllabus corpus and have no gold chunks. They are used to evaluate whether the system refuses to answer unsupported queries.

## *3.6. Automatic Evaluation Query Generation*

Evaluation queries were automatically generated using GPT-5.5 according to the query taxonomy described above. For answerable queries, predefined gold chunks were used to generate queries and assign gold labels automatically. For Out-of-scope Queries, no gold chunks were assigned.

Prompt templates instructed GPT-5.5 to derive queries only from the provided gold chunks. The templates were customized for each query type: retaining keywords for Lexical Match Queries, reducing keyword overlap for Semantic Paraphrase Queries, and requiring multiple chunks for Multi-evidence Synthesis Queries.

### *3.7. Evaluation Metrics and Experimental settings*

The experiments used the syllabus corpus of 97 courses and the automatically generated 100-query evaluation set. The LLM-only baseline was tested on 75 answerable queries without providing any syllabus information, examining whether the LLM could answer queries without any provided course information.

For RAG-based experiments, Faithfulness and Answer Relevance were introduced from RAGAs [8] to evaluate generation performance. Gold Chunk Precision@K and Gold Chunk Recall@K were designed to evaluate retrieval performance. Refusal Rate is used to evaluate the boundary capability of the lightweight LLM. Three retrieval strategies were tested under gradually increasing Top-k settings. The same local generation model and prompt were used consistently. The test consists of five metrics:

- Faithfulness evaluates how faithful the answer is to the input context, judging whether statements extracted from the answer are supported by the information present in the input context.
- Answer Relevance measures how relevant the answer is to the user's query, calculating similarity between the generated three queries and the original query.
- Gold Chunk Precision@K tests the retriever's ability to retrieve necessary chunks. It is calculated as:

$$\text{Gold Chunk Precision@K} = \frac{|Retrieved@K\ \cap\ Gold\ Chunks|}{|Retrieved@K|}$$

- Gold Chunk Recall@K tests to what extent the retriever can collect all required gold chunks. It is calculated as:

$$\text{Gold Chunk Recall@K} = \frac{|Retrieved@K \cap Gold\ Chunks|}{|Gold\ Chunks|}$$

- Refusal Rate tests LLM's ability to correctly refuse to answer out-of-scope queries. It is calculated as:

$$\text{Refusal Rate} = \frac{N_{\text{refused}}}{N_{\text{out-of-scope}}}$$

## 4. Results and Discussions

This section reports the experimental results. We first present the LLM-only baseline, then compare the three retrieval strategies across different query types and finally analyze answer generation quality and refusal behavior.

### *4.1. LLM-only Baseline*

As shown in Figure 3, the model refused to answer 70 answerable queries, resulting in a Refusal Rate of 93.3%. This result indicates that relying only on the model's internal knowledge is insufficient for answering advising queries that require external syllabus information. Therefore, introducing RAG is meaningful for academic advising.

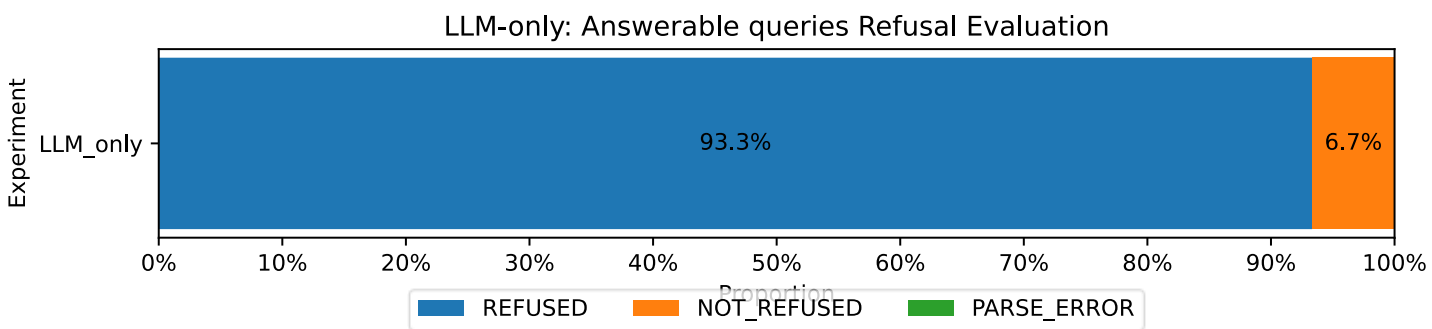

Figure 3. Refusal rate of the LLM-only baseline on 75 answerable queries without provided syllabus information.

## *4.2. Retrieval Performance across Query Types*

Figure 4 shows the overall performance across all answerable queries and demonstrates that dense retrieval performed close to hybrid retrieval but did not consistently outperform it, while BM25 showed lowest performance.

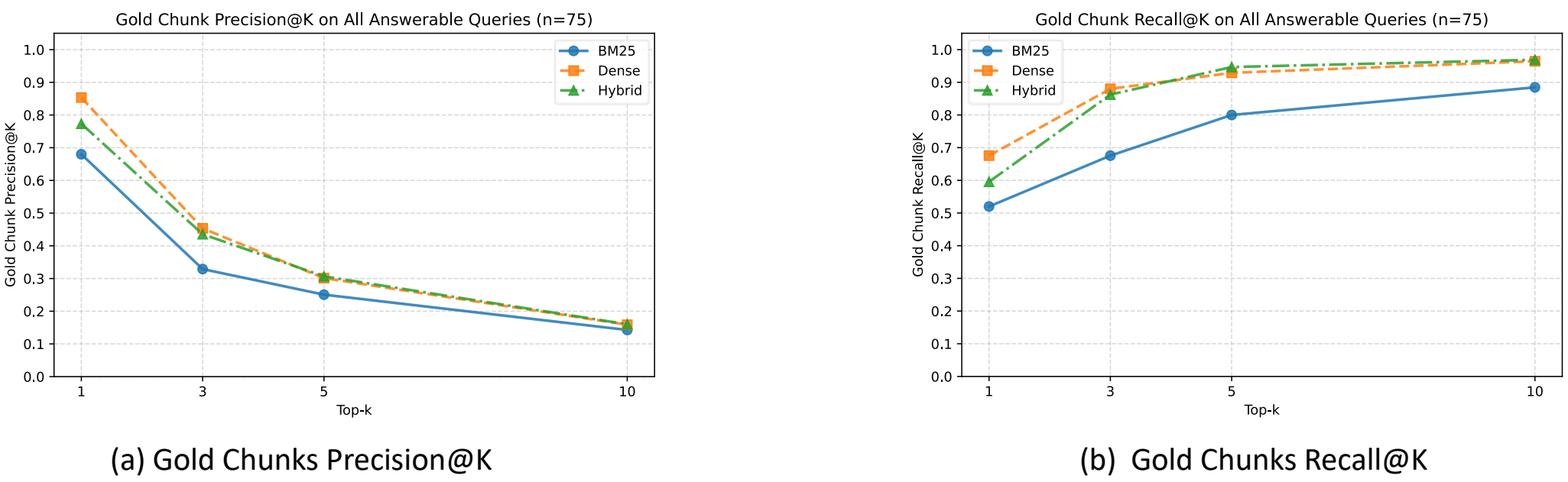

(a) Gold Chunks Precision@K (b) Gold Chunks Recall@K

Figure 4. Retrieval performance on all answerable queries.

Figure 5 presents the results for Lexical Match Queries. Because queries contain explicit keywords from one gold chunk, BM25 was expected to perform strongly at Top-1. However, the actual results show that all three retrievers achieved high Recall@K, and dense retrieval achieved the highest Precision@K. This suggests that even for keyword-overlap queries, dense retrieval can still be competitive in the syllabus corpus.

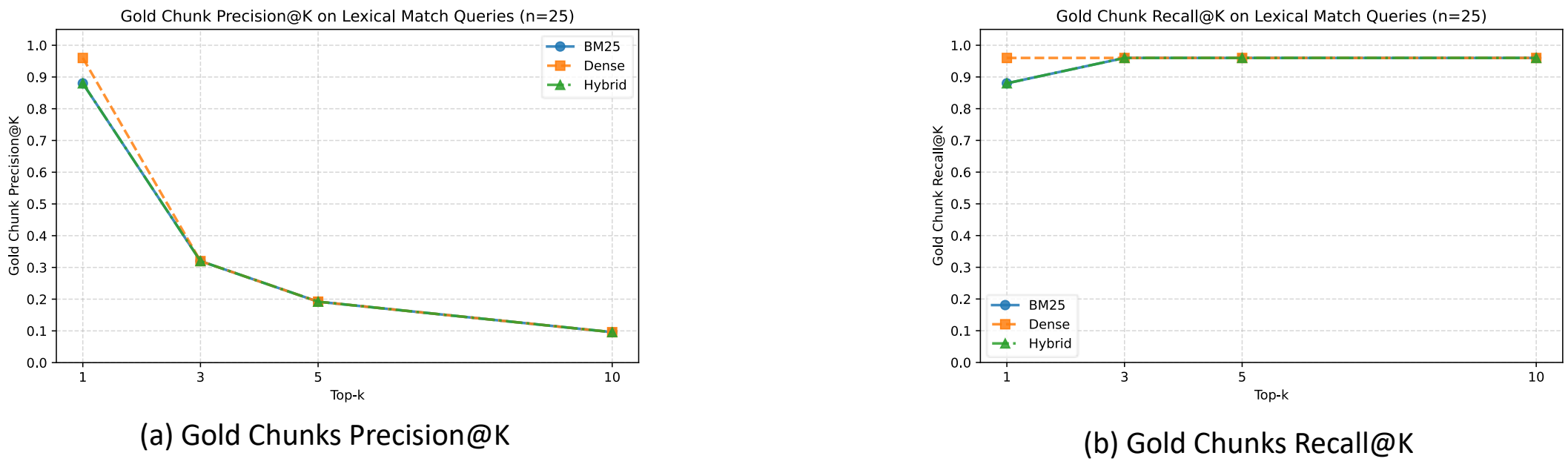

(a) Gold Chunks Precision@K (b) Gold Chunks Recall@K

Figure 5. Retrieval performance on Lexical Match Queries, which require only one gold chunk.

Figure 6 reports the results of Semantic Paraphrase Queries. Since these queries express the same meaning with reduced keyword overlap from one gold chunk, dense and hybrid retrieval were expected to outperform BM25 at Top-1. The actual results confirm this expectation, as dense and hybrid retrieval clearly outperformed BM25. This indicates that semantic matching is more effective when surface lexical overlap is reduced.

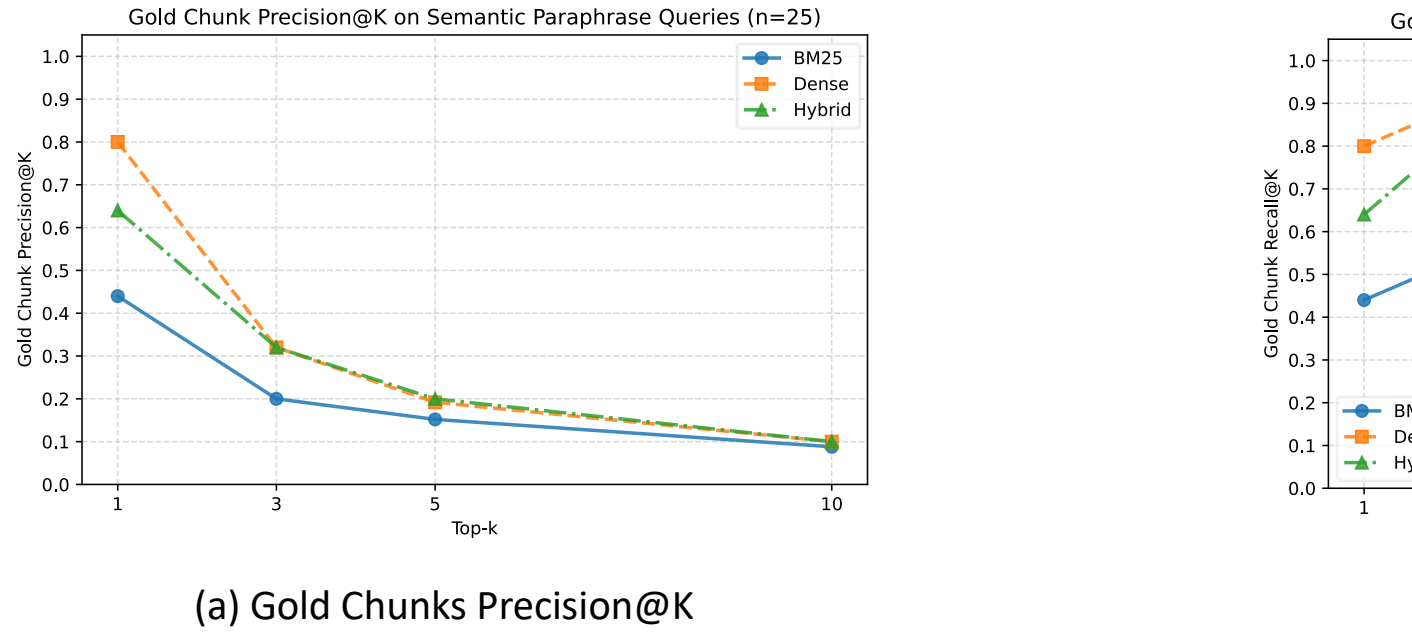

(a) Gold Chunks Precision@K

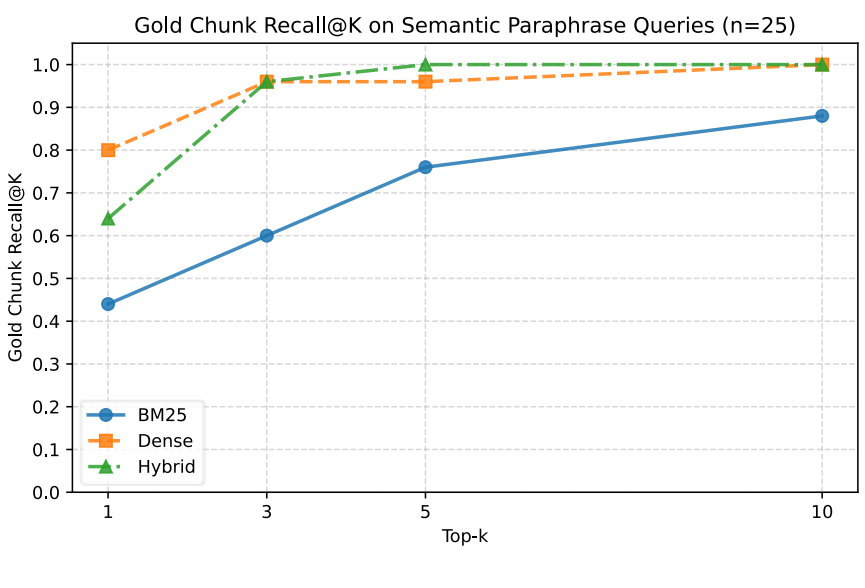

(b) Gold Chunks Recall@K

Figure 6. Retrieval performance on Semantic Paraphrase Queries, which require only one gold chunk.

Figure 7 summarizes the results for Multi-evidence Synthesis Queries. Given that hybrid retrieval combines lexical and semantic signals, hybrid retrieval was expected to outperform other retrieval strategies in this setting. However, the results show only a small difference between hybrid and dense retrieval. Both strategies achieved higher Recall@K than BM25, while BM25's Precision@K decreased more rapidly as Top-k increased. This indicates that BM25 introduced more irrelevant chunks as more results were retrieved.

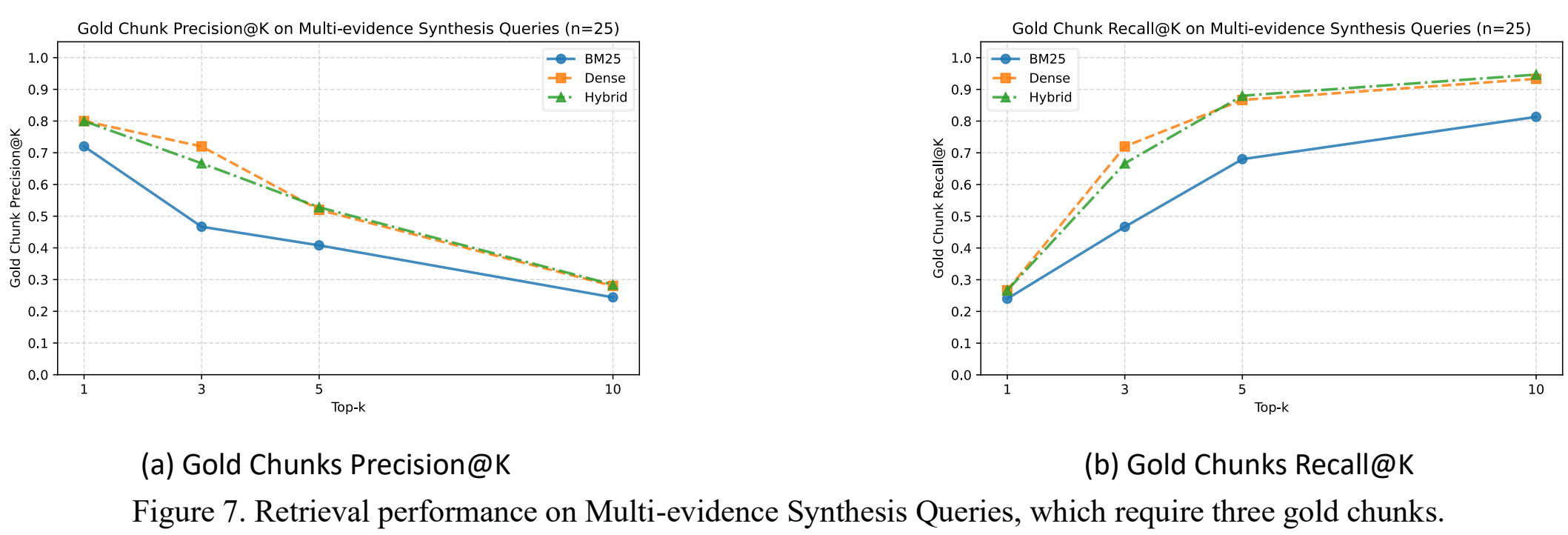

(a) Gold Chunks Precision@K (b) Gold Chunks Recall@K

Figure 7. Retrieval performance on Multi-evidence Synthesis Queries, which require three gold chunks.

These findings verified that retrieval strategies have different strengths depending on query type. They also suggest that fixed hybrid fusion did not provide a clear advantage over dense retrieval. A possible reason is that RRF fusion may also introduce non-gold chunks from other retrieval results. Therefore, hybrid retrieval may require a dynamic weighting strategy to adjust the balance between lexical and semantic signals based on retrieval complexity.

*4.3. Answer Generation Quality*

In this experimental setting, the retriever, generation model, and prompt were kept fixed, while only Top-k was varied. Therefore, Figure 8 examines how generation quality changes when the LLM receives an increasing amount of retrieved context with lower retrieval precision.

Since the generation model is a lightweight local LLM with a limited context-handling ability, answer generation quality was expected to gradually decrease as Top-k increased. The results partially verified this expectation. Faithfulness remained relatively stable at lower Top-k values and decreased after Top-10. This indicates that adding too many unnecessary chunks may contribute to generation quality decreases.

On the other hand, Answer Relevance remained stable across most Top-k settings. Answer Relevance score is based on semantic similarity between the generated answer and the original query. Therefore, a plausible but unsupported answer may still receive a high score if it is highly similar to the query. This limitation makes it difficult to determine whether a highly relevant answer answers the query correctly.

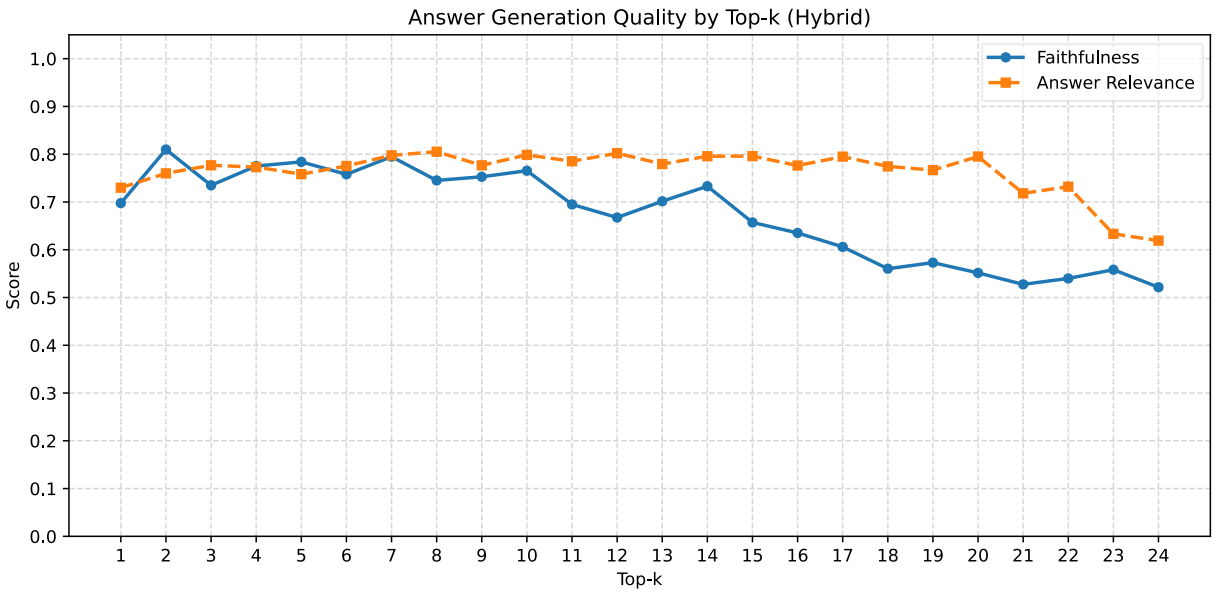

Figure 8. Generation performance for all answerable queries.

*4.4. Boundary capability*

Figure 9 demonstrates that the decline differed across retrieval strategies. These queries have no supported evidence in the syllabus corpus, so the system is expected to refuse to answer all of them. The results show that Refusal Rate generally decreased as Top-k increased. This indicates that increasing the amount of input context weakened the model's boundary capability. Although the input context did not provide any evidence for answering, the model became more likely to generate unsupported answers when more input context was provided.

The possible reason is that some input contexts may have appeared superficially related to the query, making the model more likely to generate plausible answers instead of refusing. This suggests that boundary capability is affected not only by the amount of input context, but also by the type of retrieved noise.

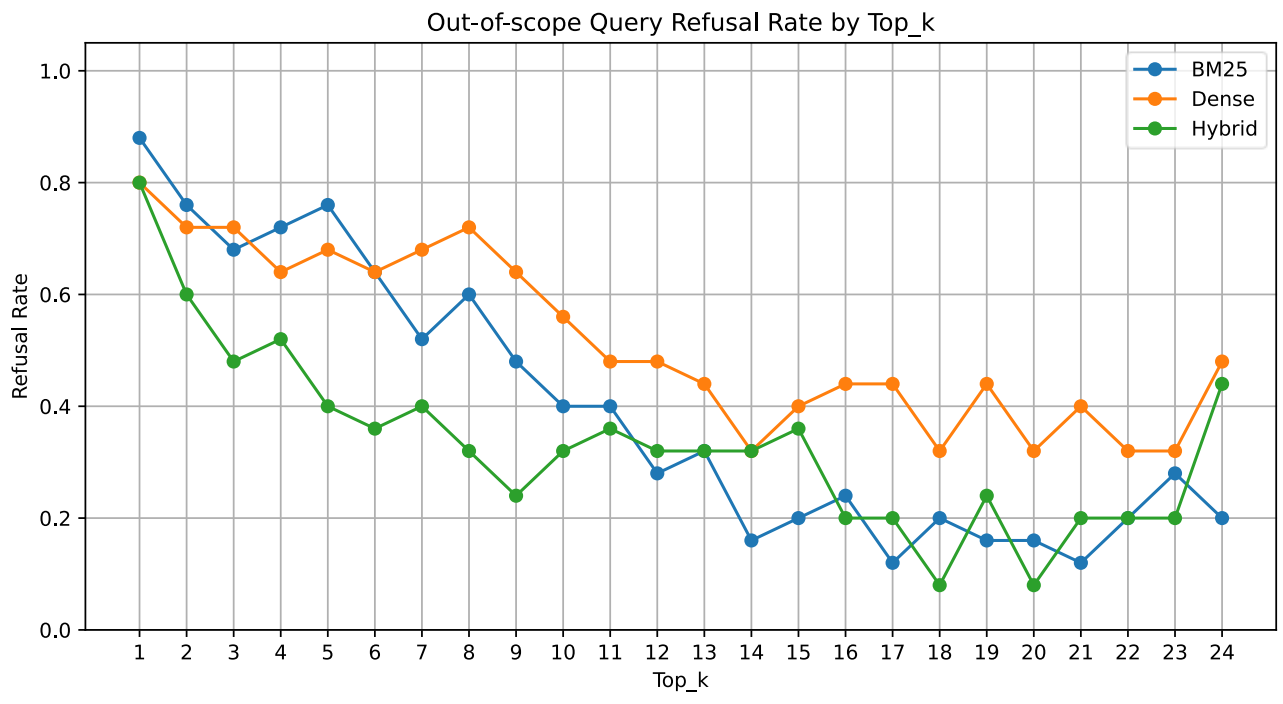

Figure 9. Refusal Rate on 25 unanswerable Out-of-scope Queries.

**5. Conclusion**

This study developed Syllabot, a locally deployed system that supports students' course selection in a privacy-conscious setting.

This study proposed a retrieval-oriented query taxonomy that defines retrieval complexity by the number of gold chunks required and the degree of lexical overlap between the query and the gold chunks. Based on this taxonomy, the study designed an automatic query generation method in which queries are generated from predefined gold chunks. This made it possible to construct an evaluation set with gold chunk labels and to examine retrieval performance under different levels of retrieval complexity.

This study also proposed Gold Chunk Precision@K and Gold Chunk Recall@K as retrieval-oriented evaluation metrics. These metrics directly evaluate whether the retriever successfully retrieves the evidence required to answer each query. Using this evaluation framework, the experiments showed that retrieval tasks with different complexity require different retrieval strategies. In this syllabus corpus, dense retrieval performed more effectively than BM25 and fixed hybrid retrieval in many settings. In addition, the boundary capability evaluation showed that increasing the amount of input context can increase the risk of hallucination in generated answers.

However, there are several limitations. Although dense retrieval performed well in this syllabus corpus, lexical retrieval still has unique advantages. In syllabus documents that contain more complex or distinctive terms, BM25 may perform more effectively because such terms can provide stronger lexical signals. Therefore, hybrid retrieval should not simply combine lexical and semantic retrieval with fixed weights. A dynamically weighted hybrid strategy may be more effective for syllabus documents.

Another limitation concerns similarity-based retrieval and evaluation. Chunks with high semantic similarity to a query do not necessarily contain valid evidence for answering it. Such related but unsupported chunks can reduce retrieval precision and increase hallucination risk. Similarly, Answer Relevance has limited generality as an answer quality metric. Because it is based mainly on semantic similarity between the generated answer and the query, an answer that appears relevant but incorrect may still receive a high score. Conversely, a correct refusal such as “I do not know” may receive a low score because it is semantically distant from the original query.

Future work should explore evidence-oriented retrieval beyond simple similarity matching. This retrieval strategy aims to identify chunks that contribute evidence for answering the query rather than chunks that are merely semantically similar. It may help retrieve gold chunks more accurately and reduce hallucination caused by superficially related but unsupported input context.

Moreover, evaluation of answer generation should incorporate evidence-based judgment. Instead of only measuring whether an answer is semantically related to the query, evaluation should examine whether the answer contributes to resolving the query and whether each statement is supported by retrieved evidence. This direction may provide a more reliable evaluation framework for RAG-based systems.

## Acknowledgements

In conducting this research, we received substantial support from Shiga University in terms of the research environment and computational sources. We would like to express sincere gratitude for this support.